\documentclass[final]{cvpr}

\usepackage{times}
\usepackage{epsfig}
\usepackage{graphicx}
\usepackage{amsmath}
\usepackage{amssymb}
\usepackage{booktabs}       
\usepackage{subcaption}
\usepackage{footnote}
\usepackage{floatrow}
\usepackage{adjustbox}
\usepackage{caption}
\usepackage{multirow}
\usepackage{xcolor}
\usepackage{dblfloatfix}
\newfloatcommand{capbtabbox}{table}[][\FBwidth]

\usepackage[pagebackref=true,breaklinks=true,colorlinks,bookmarks=false]{hyperref}

\def\cvprPaperID{48} 
\def\confYear{CVPR 2021}

\begin{document}

\title{Instance-Level Task Parameters: A Robust Multi-task Weighting Framework}

\author{Pavan Kumar Anasosalu Vasu, Shreyas Saxena, Oncel Tuzel\\
Apple Inc.\\
{\tt\small \{\href{mailto:panasosaluvasu@apple.com}{panasosaluvasu},\,\href{mailto:shreyas_saxena@apple.com}{shreyas\_saxena},\,\href{mailto:ctuzel@apple.com}{ctuzel}\}@apple.com}
}

\maketitle

\begin{abstract}
  Recent works have shown that deep neural networks benefit from multi-task learning by learning a shared representation across several related tasks. However, performance of such systems depend on relative weighting between various losses involved during training. Prior works on loss weighting schemes assume that instances are equally easy or hard for all tasks. In order to break this assumption, we let the training process dictate the optimal weighting of tasks for every instance in the dataset. More specifically, we equip every instance in the dataset with a set of learnable parameters (instance-level task parameters) where the cardinality is equal to the number of tasks learned by the model. These parameters model the weighting of each task for an instance. They are updated by gradient descent and do not require hand-crafted rules. We conduct extensive experiments on SURREAL and CityScapes datasets, for human shape and pose estimation, depth estimation and semantic segmentation tasks. In these tasks, our approach outperforms recent dynamic loss weighting approaches, e.g. reducing surface estimation errors by 8.97\% on SURREAL. When applied to datasets where one or more tasks can have noisy annotations, the proposed method learns to prioritize learning from clean labels for a given task, e.g. reducing surface estimation errors by up to 60\%. We also show that we can reliably detect corrupt labels for a given task as a by-product from learned instance-level task parameters.
\end{abstract}
\section{Introduction}

Convolutional neural networks have been successful for a wide variety of computer vision tasks. However, for complete visual 
understanding of the world a system must be able to perform multiple vision tasks simultaneously. Having independent set 
of neural networks to perform these tasks is highly inefficient especially in limited compute environments like smartphones,
wearable devices and robots. Hence, it is desirable to have a single network that can perform multiple tasks. Moreover, multi-task learning improves generalization by learning a shared representation among related tasks, seen in works like~\cite{eigen2014predicting,ren2015faster,he2017mask}. While many architectures have been proposed for multi-task learning, balancing losses from various tasks during training is equally important for model to learn shared features useful across tasks.

A standard setting is to rely on hand-crafted rules to identify a fixed weighting of tasks~\cite{ranjan2016hyperface,dvornik2017blitznet,teichmann2016multinet,kokkinos2016ubernet}.  Other works consider learning adaptive or dynamic weighting for tasks~\cite{chen2017gradnorm,liu2018endtoend,kendall2017multitask,guo2018dtp,leang2020dynamic}. These methods make a strong assumption that all samples in a dataset are equally hard for a task. 
For example, if we are training a multi-task network to perform segmentation and pose estimation, there can be images in the dataset with varying degree of self occlusion. For images with high self occlusion, the uncertainty in pose estimation task is higher compared to segmentation task. Prior task weighting schemes do not model the task uncertainty for every sample in the dataset. Recently data parameters~\cite{saxena2019dataparams} was introduced in an effort to model uncertainty of a sample as temperature value of scaled softmax function in a fully differentiable manner. But their framework does not support regression tasks nor a setup with multiple tasks.
Our method, instance-level task parameters, models the aleatoric uncertainty~\cite{kendall2017uncertainties} of tasks for every sample in the dataset in a fully differentiable manner. Our framework supports both classification and regression objectives and works in a multi-task setting. Since our method models task uncertainties for every sample in the dataset, in the presence of label corruption for one or more tasks, our method learns to prioritize learning the task from clean labels. The main contributions of our work are:
\begin{enumerate}
\item We introduce instance-level task parameters that learns dynamic weighting of tasks for every sample in the dataset. The new learning algorithm leads to better generalization across all tasks.
\item When dataset has noisy labels for some of the tasks, our method learns to prioritize learning from samples with clean labels for a given task.
\item We conducted experiments on SURREAL~\cite{surreal2017} and CityScapes~\cite{cordts2016cityscapes} datasets for human shape, human pose, depth estimation and semantic segmentation tasks.
Our method improves over recent multi-task weighting approaches, both for noise-free and noisy settings. 
\item We show that our method can accurately identify samples with corrupt labels (e.g. 99.7\% in a controlled corruption scenario).
\end{enumerate}

\section{Related Work}

Multi-task learning was investigated well before deep neural network became popular~\cite{Baxter_2000,Thrun96islearning,caruana1998multitask}. Since the advent of deep neural networks there have been many neural network architectures proposed for multi-task learning. They can be broadly classified into hard parameter sharing and soft parameter sharing architectures~\cite{ruder2017overview}. Architectures based on soft parameter sharing consists of independent network parameters for each task and a mechanism for feature sharing between tasks. Examples of such architectures include multi-task attention networks~\cite{liu2018endtoend}, neural discriminative dimensionality reduction networks~\cite{gao2018nddrcnn}, cross-stitch networks~\cite{misra2016crossstitch}, sluice networks~\cite{ruder2017latent}. These methods are not parameter efficient as every new task requires a new set of network parameters. Other parameter efficient methods like~\cite{mallya2017packnet} and \cite{mallya2018piggyback}, are trained in an iterative manner where network redundancies are exploited. Some methods have shown to learn multi-branch network architectures using greedy optimization based on task affinity measures~\cite{lu2016fullyadaptive,v2019branched} or convolutional filter grouping~\cite{bragman2019stochastic,strezoski2019task}. Hard parameter sharing is most commonly employed in multi-task learning~\cite{teichmann2016multinet,ranjan2016hyperface,dvornik2017blitznet,huang2015crossdomain,kokkinos2016ubernet,bilen2016integrated,jou2016deep}. These architectures consists of a backbone neural network that is shared across all tasks and a shallow task specific layer. In our work, we employ hard parameter sharing, see Section \ref{sec:implementation_details} for more details.

Another important area of focus in multi-task learning is optimal learning strategy. For a given network architecture, choice of task weighting strategy can influence how well the network can model different tasks. Earlier works~\cite{ranjan2016hyperface,dvornik2017blitznet,teichmann2016multinet,kokkinos2016ubernet} tuned for task weights by extensively searching for optimal task weights. More recently, adaptive methods have emerged where the pace at which various tasks are learned is modulated during the training process. In~\cite{sener2018multitask}, multi-task learning is cast as a multi-objective optimization problem, where weighting of different tasks is adaptively changed to reach a pareto optimal solution. In~\cite{kendall2017multitask}, homoscedastic uncertainty or task-dependent uncertainty is used to weight single-task losses. In~\cite{guo2018dtp}, harder tasks are prioritized over easy tasks in early phases of training and the hardness of a task is measured by its key performance indicator (KPI). Choice of optimal KPI for a task is not straightforward. In~\cite{chen2017gradnorm}, all tasks are encouraged to learn at a similar pace by balancing the magnitude of gradients across tasks. Instead of using gradient statistics to weight tasks appropriately, in~\cite{liu2018endtoend}, the method utilizes rate of change of task specific losses. This way it is faster to compute than~\cite{chen2017gradnorm}, but this strategy requires losses to be balanced in magnitude beforehand. More recently, meta-learning mechanisms have been explored to find optimal task weighting in \cite{leang2020dynamic}, but a major drawback of meta-learning approaches is the increased computational cost from several training runs. Perhaps the most relevant work to our current work is data parameters introduced in \cite{saxena2019dataparams}, where every sample has a class and instance parameter. However these parameters were combined to learn an optimal temperature scaling for classification task with no support for regression. Moreover their framework has no support for multiple tasks. In our formulation, we cater to both regression and classification tasks in a multi-task setting. 
\section{Instance-Level Task Parameters}
The main goal of our work is to model the importance of an instance for 
learning different tasks in a multi-task training setup. Instead of solving for a discrete assignment of an instance to a task, 
we represent the importance of an instance for a task via instance-level task parameters, where these parameters are learned along with 
the model parameters. 
To this end, we derive the multi-task loss function based on minimizing the negative log-likelihood loss as described in~\cite{kendall2017multitask}, 
but instead of modeling homoscedastic uncertainty which remain constant over all instances and varies between tasks,
we learn instance-level task parameters that model uncertainty which varies across tasks and instances. These parameters capture the task uncertainty for every instance, as different instances can have different levels of uncertainty for each task.


To formalize our intuition, let $\left(\mathbf{x}_i, \mathbf{y}_i^k\right)$ denote a training data point, where $\mathbf{x}_i$ is the input and 
$\mathbf{y}_i^k$ is the regression or classification target for the $k^{th}$ task. 
Let $\boldsymbol{\sigma} \in \mathbb{R}^{N\times K}$ denote instance-level task parameters, where $N$ is the total number of samples in training dataset and $K$ is the total number of tasks learned by the model. $\sigma_{i}^{k}$ denotes instance-level parameter of the $i^{th}$ sample, for the $k^{th}$ task. Let $L^k_i(\boldsymbol{\theta}, \sigma_i^{k})$ represent the loss function for the $k^{th}$ task on sample $i$.
In our framework for multi-task learning, we solve the following optimization problem: 
\begin{equation} 
\min_{\boldsymbol{\theta}, \boldsymbol{\sigma}} \frac{1}{N} \sum_{i=1}^N \sum_{k=1}^K L^k_i(\boldsymbol{\theta}, \sigma_i^k)
\label{eq:optimization}
\end{equation}
where in addition to the model parameters, $\boldsymbol{\theta}$, we also solve for instance-level task parameters $\boldsymbol{\sigma}$. 
The loss function $L_i^k$  could either be a classification loss or a regression loss. To simplify the presentation, throughout the paper, we drop the index $k$ from the equations when it's clear from the context.

\textbf{Regression loss}:
Given a sample $\mathbf{x}_i$, let the model $f_{\boldsymbol{\theta}}\left(\mathbf{x}_i \right)$ output the mean of a Normal distribution with variance $\sigma_i^2$, $p(\mathbf{y} | f_{\boldsymbol{\theta}}\left(\mathbf{x}_i \right), \sigma_i) = \mathcal{N}\left( f_{\boldsymbol{\theta}}\left(\mathbf{x}_i \right), \sigma_i^2 \right)$. The negative log-likelihood loss is then given by
\begin{equation}
\begin{split}
L^{reg}_i(\boldsymbol{\theta}, \sigma_i) &= -\log(p(\mathbf{y}_i | f_{\boldsymbol{\theta}}\left(\mathbf{x}_i \right), \sigma_i)) \\
& \propto \frac{\|f_{\boldsymbol{\theta}}\left(\mathbf{x}_i\right)-\mathbf{y}_i\|^2}{2{\sigma_{i}}^2} + \log({\sigma_{i}}^2) \label{eq:reg_expanded_NLL}
\end{split}
\end{equation}
Here we assume that every sample $i$ has a different variance $\sigma_i^2$. This should not be confused with heteroscedastic uncertainty as defined in~\cite{kendall2017what} where $\sigma_i$ is a function of $\mathbf{x}_i$. In our formulation $\sigma_i$ depends on both $\mathbf{x}_i$ and $\mathbf{y}_i$, modeled explicitly, hence can model the per sample noise. We do not predict it from the network but learn during training. 

\textbf{Classification loss}:
We model the classification likelihood as a scaled softmax function, where the model predictions, $f_{\boldsymbol{\theta}}\left(\mathbf{x}_i \right)$, are scaled by a positive scalar $\sigma_{i}$, $p(\mathbf{y}_i | f_{\boldsymbol{\theta}}\left(\mathbf{x}_i \right), \sigma_i) = \frac{\exp(f_{\boldsymbol{\theta}}^{y_i}\left(\mathbf{x}_i \right)/ {\sigma_{i}}^2)} {\sum_{j} \exp(f_{\boldsymbol{\theta}}^j\left(\mathbf{x}_i \right)/{\sigma_{i}}^2)}$. 

The $f_{\boldsymbol{\theta}}^{y_i}\left(\mathbf{x}_i \right)$ and $f_{\boldsymbol{\theta}}^{j}\left(\mathbf{x}_i \right)$ are the logit of the target class ${y_i}$ and logit of class $j$ respectively.  The scale, $\sigma_i$, decides how peaky or flat the discrete distribution is, thus quantifying the uncertainty on 
the $i^{th}$ data point. The negative log-likelihood is then given by
\begin{multline}
-\log(p(\mathbf{y}_i | f_{\boldsymbol{\theta}}\left(\mathbf{x}_i \right), \sigma_i)) \\
= \log \left[ \sum_{j}\exp\left(\frac{f_{\boldsymbol{\theta}}^{j}\left(\mathbf{x}_i \right)}{{\sigma_{i}}^2}\right) \right] -\frac{f_{\boldsymbol{\theta}}^{y_i}\left(\mathbf{x}_i \right)} {{\sigma_{i}}^2} \label{eq:cls_expanded_NLL}
\end{multline}

We use the approximation introduced in~\cite{kendall2017multitask} for equation (\ref{eq:cls_expanded_NLL}) and define the approximate negative log-likelihood loss function:

\begin{equation}
L^{cls}_i(\boldsymbol{\theta}, \sigma_i) = \frac{1}{{\sigma_{i}}^2} \log\left(\frac{\exp(f_{\boldsymbol{\theta}}^{y_i}\left(\mathbf{x}_i \right))} {\sum_{j} \exp(f_{\boldsymbol{\theta}}^{j}\left(\mathbf{x}_i \right))}\right) + \log({\sigma_{i}}^2)
\end{equation}

\textbf{Multi-Task loss}:
We assume that the joint multi-task likelihood factorizes over the tasks $p(\mathbf{y}_{i}^1, ..., \mathbf{y}_{i}^K|f_{\boldsymbol{\theta}}(\mathbf{x}_i), \boldsymbol{\sigma}) = \prod_{k=1}^{K} p(\mathbf{y}_{i}^k|f_{\boldsymbol{\theta}}(\mathbf{x}_i), \sigma_i^k)$ where the negative log-likelihood gives us the multi-task loss in (\ref{eq:optimization}),
$L(\boldsymbol{\theta}, \boldsymbol{\sigma}) =  \frac{1}{N} \sum_{i=1}^N \sum_{k=1}^K L^k_i(\boldsymbol{\theta}, \sigma_i^k)$.

\subsection{Optimization of Instance-Level Task Parameters} \label{sec:opt_sparse_sgd}

For illustration purposes, let us consider a multi-task loss with a regression and a classification objective:
\begin{equation} 
L(\boldsymbol{\theta}, \boldsymbol{\sigma}) = \frac{1}{N} \sum_{i=1}^N L^{reg}_i(\boldsymbol{\theta}, \sigma_i^{reg}) + L^{cls}_i(\boldsymbol{\theta}, \sigma_i^{cls})
\end{equation}
\begin{multline} 
= \frac{1}{N} \sum_{i=1}^{N} \frac{l_i^{reg}}{2(\sigma_i^{reg})^2} + \frac{l_i^{cls}}{(\sigma_i^{cls})^2} \\
+ \log\left((\sigma_i^{reg})^2\right) + \log\left((\sigma_i^{cls})^2\right) \label{eq:mtl_reg_cls}
\end{multline} 
where $l_i^{reg} = \|f_{\boldsymbol{\theta}}\left(\mathbf{x}_i\right)-\mathbf{y}_i\|^2$ is the squared $l_2$ error and $l_i^{cls} = \log\left(\frac{\exp(f_{\boldsymbol{\theta}}^{y_i}\left(\mathbf{x}_i \right))} {\sum_{j} \exp(f_{\boldsymbol{\theta}}^{j}\left(\mathbf{x}_i \right))}\right)$ is the  cross entropy error. Gradients for model parameters are computed as,
$\frac{\partial{L(\boldsymbol{\theta},\boldsymbol{\sigma})}}{\partial\boldsymbol{\theta}} = \frac{1}{N} \sum_{i=1}^{N} \left( \frac{1}{2(\sigma_i^{reg})^2} \frac{\partial{l_i^{reg}}} {\partial\boldsymbol{\theta}} + \frac{1}{(\sigma_i^{cls})^2} \frac{\partial{l_i^{cls}}} {\partial\boldsymbol{\theta}} \right)$.
When the instance-level task parameter is high for an instance, the gradient for model parameters from that instance for that task diminishes. This results in slowing down learning from that instance for that task.
Gradients for classification and regression task parameters are computed as,
$\frac{\partial{L(\boldsymbol{\theta},\boldsymbol{\sigma})}}{\partial\sigma_i^{cls}} = \frac{2}{\sigma_i^{cls}} \left(1 - \frac{l_i^{cls}}{(\sigma_i^{cls})^2} \right)$ and
$\frac{\partial{L(\boldsymbol{\theta},\boldsymbol{\sigma})}}{\partial\sigma_i^{reg}} = \frac{1}{\sigma_i^{reg}} \left(2 - \frac{l_i^{reg}}{(\sigma_i^{reg})^2} \right)$, respectively.

From the above equations we observe that, during stochastic optimization, the gradients for model parameters are averaged over $N$ samples within a batch, whereas instance parameters collect their gradients from individual samples (when sampled in a batch). 
The gradient for an instance parameter of a task is only influenced by the performance of the model for that task. It's worth noting that this is a major difference between the presented formulation and~\cite{kendall2017multitask}. 
By construction, our formulation breaks the assumption that all instances are equally hard for a task and the training process can dictate how the loss is balanced for every instance in the training dataset. 
In real world datasets where one or more tasks can have noisy annotations, the model has the freedom to choose to learn from tasks which have clean annotations for any given instance (see Section \ref{sec:learn_noisy} for more details). In~\cite{kendall2017multitask}, all samples contribute equally to the optimization regardless of their difficulty or uncertainty. 

In our experiments, we use Stochastic Gradient Descent (SGD) with momentum optimization for instance-level task parameters. We implemented sparse SGD updates, where we prevent updating instance-level task parameters based on momentum when there is no gradient (the sample is not within the current batch). Note that, the instance-level task parameters are not part of the model parameters, they are not used post training during inference, hence they do not change the model capacity.

\section{Experiments}
We conduct experiments on SURREAL~\cite{surreal2017} and CityScapes~\cite{cordts2016cityscapes}.
SURREAL is a large-scale synthetic dataset of textured humans.
The dataset contains ground truth labels for segmentation, 2D/3D pose, and SMPL~\cite{smpl2015} body parameters 
for shape estimation. CityScapes dataset consists of high resolution street-view images. 
The dataset provides pixel-wise labels for semantic segmentation and depth estimation.

\begin{figure*}[!t]
   \captionsetup{font=small}
     \ffigbox{\includegraphics[scale=0.363]{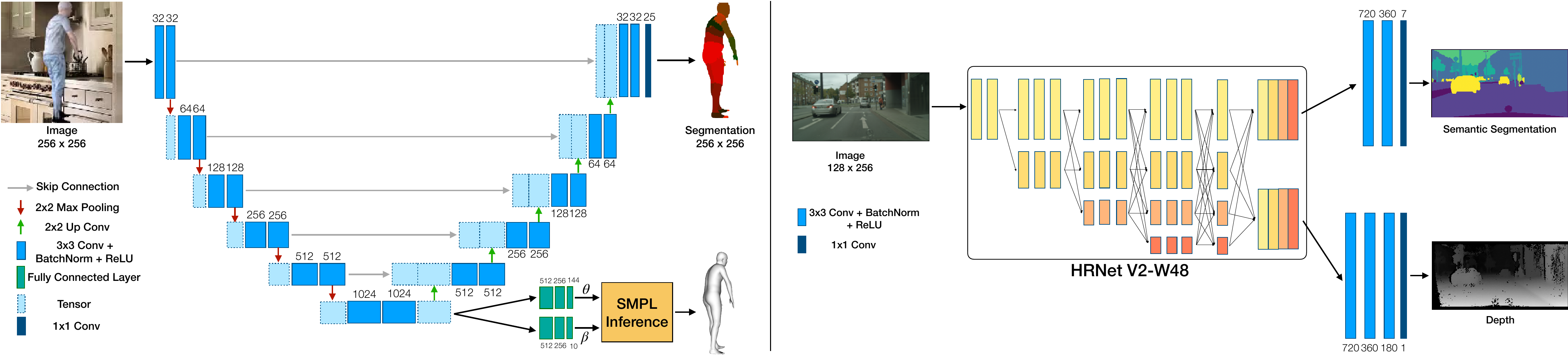}}
     {
     \caption{Architecture details \textbf{Left}: Multi-task UNet for human shape and pose estimation, \textbf{Right}: High resolution network \cite{sun2019highresolution} (HRNetV2-W48) for pixel-wise depth estimation and semantic segmentation.} \label{fig:arch_overview}
     }
\end{figure*}

\textbf{Baselines}: We compare our method against the following recently proposed loss weighting schemes. The following methods will be referenced by their acronyms in parenthesis in subsequent sections.
{
\begin{itemize}
  \itemsep-0.2em
  \item Equal Weighting (Eq. Weight): All the losses are weighted equally.
  \item MultiTask Uncertainty~\cite{kendall2017multitask} (MTU): Losses are scaled by homoscedastic uncertainty of the respective tasks.
  \item Dynamic Weight Averaging~\cite{liu2018endtoend} (DWA): Loss weighting is based on the rate of change of loss computed for each task, which is normalized using a softmax function. Losses have to be balanced for this scheme to work well. Apart from hyperparameters to balance losses, this scheme also requires an optimal temperature for softmax function. In Section \ref{sec:implementation_details}, we will discuss them in more detail. 
  \item Geometric Loss Strategy~\cite{chennupati2019multinet} (GLS): The total loss is geometric mean of individual task losses.
\end{itemize}
}

\subsection{Implementation Details} \label{sec:implementation_details}
 To have a fair comparison with baseline loss weighting methods, we used the same neural network architecture, optimizer and learning rate (for model parameters) for all the methods. The following are the remaining hyperparameters pertaining multi-task weighing. Equal weighting and GLS~\cite{chennupati2019multinet} do not have any hyperparameters.
 For task parameters in MTU~\cite{kendall2017multitask}, we use the optimizer and learning rate as prescribed in the original paper. DWA~\cite{liu2018endtoend} obtains good results only when the losses are balanced in magnitude before computing its rate of change. Other works like~\cite{v2020revisiting}, have also made a note of this. Loss scaling factors for each of the models for DWA is reported in the respective sections. They are set by a grid search to find the optimal temperature for scaled softmax function used to compute the loss weights. Note, results reported for these loss weighting schemes are from our re-implementation of the respective methods.

Optimizing instance-level task parameters $\boldsymbol{\sigma}$ with gradient descent requires constraint optimization 
with constraint $\boldsymbol{\sigma} \geq 0$. Instead, we use log parameterization $\log(\sigma^2)$, which allows unconstrained optimization and can be mapped back using exponential mapping. Moreover, we clamp $\log(\sigma^2)$  within the range -4.0 to 4.0. For all experiments, instance-level task parameters, $\log(\sigma^2)$, are initialized as $0$.
For our method, Instance-Level Task Parameters (ILT Params), the only hyperparameter is the learning rate for the optimizer (sparse SGD with momentum, described in Section \ref{sec:opt_sparse_sgd}). We specify the value of this parameter in each experiment section.  We report mean and standard deviation of all the metrics across 3 runs for all the loss weighting methods.

\subsection{Human Shape and Pose Estimation}
In this section, we evaluate our method on a model performing semantic segmentation, regression of shape and 3D pose parameters
of a parametric human shape model, SMPL~\cite{smpl2015}. We preprocess SURREAL~\cite{surreal2017} dataset, by cropping and resizing images to resolution $[256\times256]$ and use the augmentation schemes as described in BodyNet~\cite{varol2018bodynet}. We use UNet architecture introduced in~\cite{ronneberger2015unet} as the backbone and attach task specific layers to the decoder. This architecture is referenced as Multi-Task UNet. 
The segmentation head has 25 channels for the 25 classes prescribed by SURREAL~\cite{surreal2017} dataset. From the bottleneck layer 
we add two branches of 3 fully-connected layers to regress shape $\boldsymbol{\beta} \in \mathcal{R}^{10}$ 
and pose parameters $\boldsymbol{\Theta} \in \mathcal{R}^{144}$ of SMPL. We regress 6D rotations prescribed in~\cite{zhou2018continuity} 
for all 24 joints of SMPL. We also incorporate loss on the vertex reconstructions by implementing SMPL as a differentiable layer as described in~\cite{kanazawa2017endtoend}. Implementation details are discussed in detail in supplementary material, Section A.1. The dataset provides annotations for all the tasks listed above, hence we have the following
losses, $l_i^{cls}$, pixel-wise cross entropy loss for data point $i$; 
%
%
$l_i^{\beta}$, $l_i^{\Theta}$ and $l_i^{3d}$ are squared $L2$ losses on shape, pose parameters and vertex
reconstruction for data point $i$. As there are four objectives to optimize over, we instantiate four parameters per instance, 
$\sigma_i^{cls}$, $\sigma_i^{\Theta}$, $\sigma_i^{\beta}$ and $\sigma_i^{3d}$. We set the learning rate for instance-level task parameters as 2.0. For DWA~\cite{liu2018endtoend}, $l_i^{cls}$, $l_i^{\beta}$, $l_i^{\Theta}$ and $l_i^{3d}$ are scaled by a factor of 0.4, 0.05, 1.0 and 1.0 respectively and setting softmax temperature to 0.5 produced the best results.
Table \ref{tab:hsm_clean} shows results on SURREAL test dataset.
ILT Params is the best performing method on all tasks. We obtain 4.78\% reduction in surface errors compared to BodyNet~\cite{varol2018bodynet}. Compared to the best performing multi-task weighting method, DWA~\cite{liu2018endtoend}, ILT Params obtains 8.97\% reduction in surface errors.

\begin{table}[!t]
\captionsetup{font=small}
\begin{adjustbox}{width=8.3cm}
\begin{tabular}{cccc} 
\toprule
Method & Surface Errors & MPJPE & Pixel Accuracy   \\ 
\midrule
Pavlakos \textit{et al.}~\cite{pavlakos2018learning}* & 151.5 & - & -\\
HMNet~\cite{venkat2019humanmeshnet}* & 86.6 & 71.9 & -\\
SMPLR~\cite{madadi2018smplr}* & 75.4 & 55.8 & - \\
Tung \textit{et al.}~\cite{tung2017selfsupervised}* & 74.5 & 64.4 & - \\
BodyNet (single-view)~\cite{varol2018bodynet}* & 67.7 & - & -\\ 
BodyNet (multi-view)~\cite{varol2018bodynet}* & 65.8 & - & -\\ 
\midrule
\midrule
Eq. Weight & 75.15 $\pm$ 0.33 & 65.11 $\pm$ 0.44 & 98.03 $\pm$ 0.014 \\
MTU~\cite{kendall2017multitask} & 70.69 $\pm$ 0.64 & 61.04 $\pm$ 0.64 & 98.09 $\pm$ 0.007\\
DWA~\cite{liu2018endtoend} & 68.83 $\pm$ 0.77 & 59.14 $\pm$ 0.67 & 98.14 $\pm$ 0.006 \\
GLS~\cite{chennupati2019multinet} & 69.63 $\pm$ 1.56 & 59.96 $\pm$ 1.49 & 98.10 $\pm$ 0.020 \\
ILT Params (Ours) & \textbf{62.65} $\pm$ \textbf{0.16} & \textbf{53.49} $\pm$ \textbf{0.13} & \textbf{98.25} $\pm$ \textbf{0.001}\\ 
\bottomrule
\end{tabular}
\end{adjustbox}
{
  \caption{Comparison with state-of-the-art methods and baseline loss weighting methods on SURREAL test dataset. Both Surface Errors and MPJPE are in \textit{mm} resolution and accuracy is in percentage. Results for various loss weighting methods are from our re-implementation of the respective methods. (*) denotes results cited from the original paper. All metrics reported in lower half of the table are using Multi-task UNet architecture. For pixel accuracy metric, higher is better, and for surface errors and MPJPE metrics, lower is better.} \label{tab:hsm_clean}
}
\end{table}

\subsection{Pixel-wise Depth Estimation and Semantic Segmentation}
In this section, we conduct experiments on joint pixel-wise depth regression and semantic segmentation tasks on CityScapes dataset~\cite{cordts2016cityscapes}. We closely follow the setup described in~\cite{liu2018endtoend}, which uses 7-class version of semantic labels and paired inverse depth labels. All images and labels are resized to $[128\times256]$. We use HRNetV2~\cite{sun2019highresolution} backbone for all the methods, which is a stronger backbone than SegNet~\cite{badrinarayanan2015segnet} used in~\cite{liu2018endtoend}. Implementation details are discussed in detail in supplementary material, Section A.2. 
We set the learning rate for instance-level task parameters to 4.0. For DWA~\cite{liu2018endtoend}, both losses are scaled by 1.0 and softmax temperature is set to 2.0 as described in~\cite{liu2018endtoend}. 
Table \ref{tab:cs_clean} shows results on CityScapes validation dataset.  
Unlike other loss weighting methods where one or more tasks are not learned to satisfaction, our method is either the best or (closely following) second best method for each task.

\begin{table}[!t]
\captionsetup{font=small}
  \begin{adjustbox}{width=8.3cm}
  \begin{tabular}{ccccc} 
	\toprule
    \multirow{2}{*}[-3pt]{Method}  & \multicolumn{2}{c}{Segmentation} & \multirow{2}{*}[5.3pt]{Depth} \\ 
    \cmidrule{2-4} &  mIoU & Pix Acc & Abs Err\\ 
    \midrule
	Eq. Weight & 63.73 $\pm$ 0.16 & 94.48 $\pm$ 0.04 & 0.0115 $\pm$ 0.0001 \\
	MTU~\cite{kendall2017multitask} & 62.63 $\pm$ 0.20 & 94.01 $\pm$ 0.14 & 0.0113 $\pm$ 0.0003 \\
	DWA~\cite{liu2018endtoend} & 63.76 $\pm$ 0.13 & 94.49 $\pm$ 0.02 & 0.0117 $\pm$ 0.0001 \\
	GLS~\cite{chennupati2019multinet} & 61.85 $\pm$ 0.37 & 94.04 $\pm$ 0.12 & \textbf{0.0111 $\pm$ 0.0003} \\
	ILT Params (Ours) & \textbf{63.98 $\pm$ 0.22} & \textbf{94.60 $\pm$ 0.04} & 0.0112 $\pm$ 0.0001 \\
	\bottomrule
  \end{tabular}
  \end{adjustbox}
 {
  \caption{Comparison with baseline loss weighting methods on CityScapes validation dataset. For all the experiments we use HRNetV2 backbone~\cite{sun2019highresolution}. For segmentation metrics higher is better and for depth metrics lower is better.} \label{tab:cs_clean}
}
\end{table}

\begin{figure*}[!t]
   \captionsetup{font=small}
     \ffigbox{\includegraphics[scale=0.162]{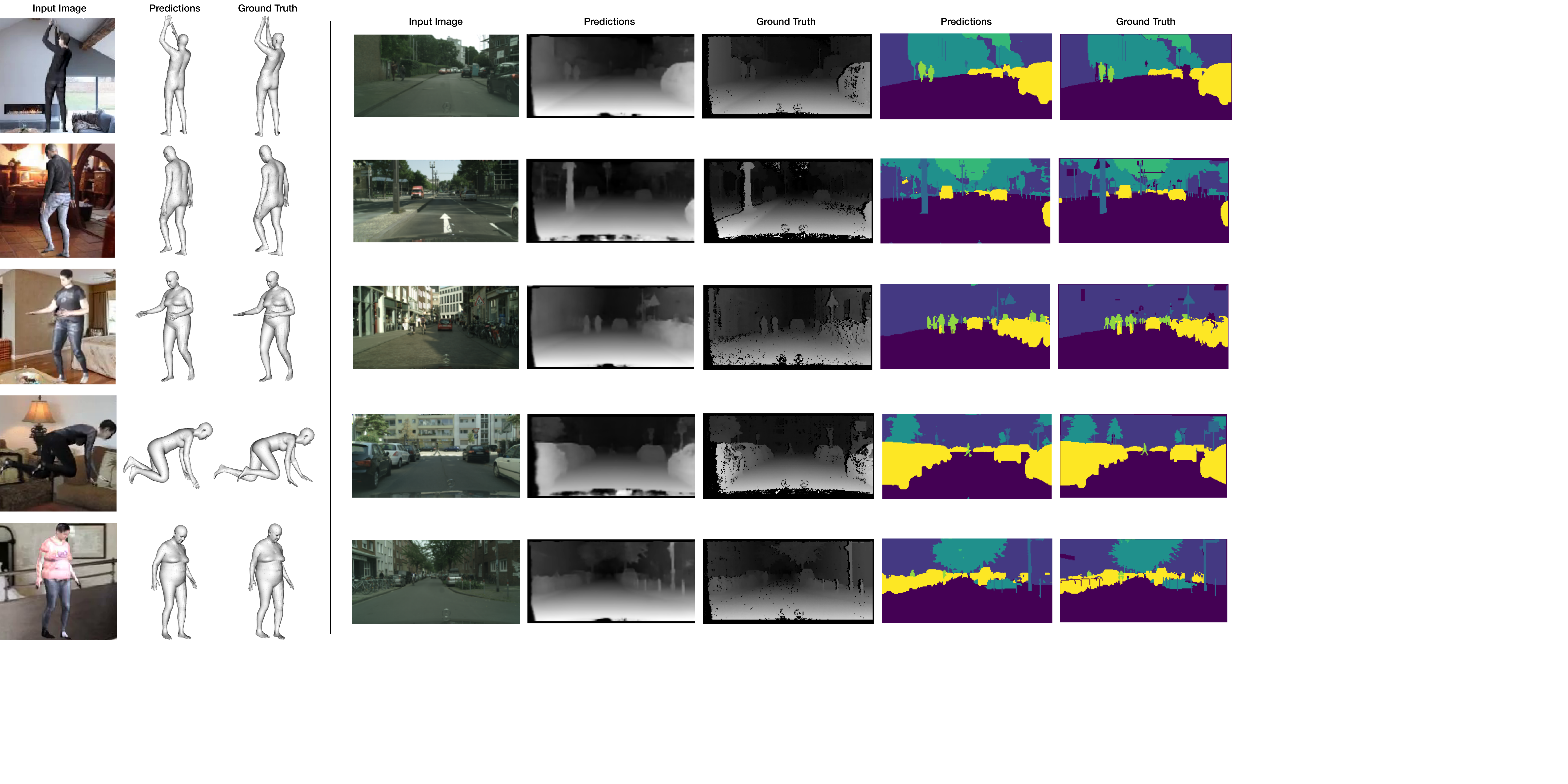}}
     {
     \caption{\textbf{Left}: Qualitative results on human pose and shape estimation from Multi-task UNet trained using our method on SURREAL test dataset, \textbf{Right}: Qualitative results on semantic segmentation and depth estimation from HRNetV2-W48 trained using our method on CityScapes validation dataset.} \label{fig:qualitative_analysis}
     }
\end{figure*}

\subsection{Computational Requirement} \label{sec:computational_requirements}
The memory and compute overhead of the method is quite small. In our experiments on SURREAL (larger dataset than CityScapes), there are 400k instance-level task parameters on top of 40 million model parameters (~1\% overhead). Compute overhead is negligible as only those parameters pertaining to instances sampled in mini-batch are updated at every step, on average both instance-level task parameters experiments and baseline experiments took 28 minutes per epoch. Note that as the dataset size scales, the computational overhead stays constant, where as memory overhead scales linearly. For very large datasets, instance-level task parameters can be instantiated on CPU instead of GPU and only those parameters pertaining to samples of a mini-batch can be loaded to GPU to be updated at every step. This design allows GPU RAM to be fully utilized by model parameters.

\begin{table*}[!b]
\captionsetup{font=small}
\noindent
\begin{floatrow}\CenterFloatBoxes
\capbtabbox{
  \begin{adjustbox}{width=7.75cm}
  \begin{tabular}{cccc} 
    \toprule 
    Method & Surface Errors & MPJPE & Pixel Accuracy \\
    \midrule
    Eq. Weight & 77.38 $\pm$ 0.71 & 67.19 $\pm$ 0.75 & 97.59 $\pm$ 0.014 \\
    MTU~\cite{kendall2017multitask} & 77.44 $\pm$ 0.39 & 67.37 $\pm$ 0.27 & 97.50 $\pm$ 0.033\\
    DWA~\cite{liu2018endtoend} & 74.02 $\pm$ 0.09 & 63.90 $\pm$ 0.08 & 97.04 $\pm$ 0.301 \\
    GLS~\cite{chennupati2019multinet} & 77.99 $\pm$ 0.93 & 67.66 $\pm$ 0.78 & 97.49 $\pm$ 0.009 \\
    ILT Params (Ours) & \textbf{65.06 $\pm$ 0.12} & \textbf{55.62 $\pm$ 0.12} & \textbf{98.07 $\pm$ 0.009} \\ 
    \bottomrule
    \end{tabular}
  \end{adjustbox}
 }{
  \caption{Comparison with baseline loss weighting methods on SURREAL test dataset when 40\% of training dataset has corrupt segmentation labels.} \label{tab:hsm_corruption_exps_segm}
}
\capbtabbox{
  \begin{adjustbox}{width=7.75cm}
  \begin{tabular}{cccc} 
    \toprule 
    Method & Surface Errors & MPJPE & Pixel Accuracy \\
    \midrule
    Eq. Weight & 193.91 $\pm$ 8.72 & 172.025 $\pm$ 8.24 & 97.83 $\pm$ 0.007 \\
    MTU~\cite{kendall2017multitask} & 164.80 $\pm$ 16.32 & 140.80 $\pm$ 14.21 & 98.15 $\pm$ 0.006 \\
    DWA~\cite{liu2018endtoend} & 203.87 $\pm$ 1.62 & 180.01 $\pm$ 1.44 & 98.01 $\pm$ 0.017 \\
    GLS~\cite{chennupati2019multinet} & 195.14 $\pm$ 11.47 & 168.65 $\pm$ 10.64 & 98.17 $\pm$ 0.005 \\
    ILT Params (Ours) & \textbf{66.19 $\pm$ 0.28} & \textbf{56.82 $\pm$ 0.27} & \textbf{98.27 $\pm$ 0.006} \\ 
    \bottomrule
    \end{tabular}
  \end{adjustbox}
 }{
  \caption{Comparison with baseline loss weighting methods on SURREAL test dataset when 40\% of training dataset has corrupt pose and shape labels.} \label{tab:hsm_corruption_exps_pose}
}
\end{floatrow}
\end{table*}

\subsection{Analysis of Instance-Level Task Parameters} \label{sec:curriculum_analysis}
\begin{figure*}[!htb]
   \captionsetup{font=small}
   \ffigbox{\includegraphics[scale=0.355]{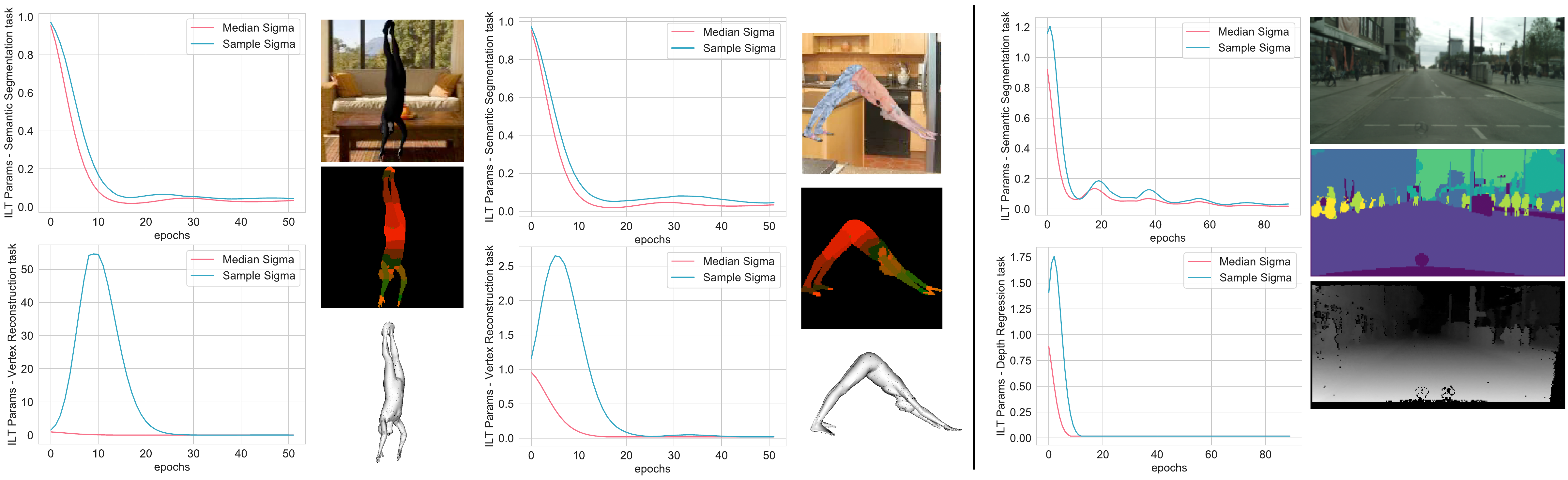}}
     {
     \caption{\textbf{Left}: Analysis of instance-level task parameters for two samples over the course of training human shape and pose estimation model on SURREAL, \textbf{Right}: Analysis of instance-level task parameters for a sample over the course of training semantic segmentation and depth regression model on CityScapes.} \label{fig:sigma_analysis} \label{fig:cirriculum_examples}
     }
\end{figure*}

To understand how instance-level task parameters evolve during the course of training, we pick a couple of images and plot their task parameters, for reference, we plot median of the respective task parameters over the entire dataset. In left plots of Figure \ref{fig:cirriculum_examples}, we show couple of examples where it is easy to perform semantic segmentation but not vertex reconstruction, as the human is in an unusual pose. During the initial stages of learning, $\sigma^{3d}$ increases which delays learning the difficult vertex reconstruction task. Once easier poses are learned, after 10$^{th}$ epoch, $\sigma^{3d}$ decreases and the model starts learning to perform the vertex reconstruction task from these images. Conversely, since the semantic segmentation task is relatively easy, the neural network starts learning segmentation task for these images from the beginning which is shown by decreasing $\sigma^{cls}$ parameter from the first epoch.
Similarly, in right plot of Figure \ref{fig:cirriculum_examples} we notice that depth regression task is learned at a slower pace from the given image at the beginning, compared to the semantic segmentation task as shown by instance-level task parameters. This is due to the image being difficult to reason about depth since the scene contains many foreground objects at different distances. 
It is important to note that instance-level task parameters are learnt in an online manner and adapts to state of model during optimization. From examples in Figure \ref{fig:cirriculum_examples}, it is evident that our method learns to delay learning from hard samples, but eventually reduces variance on these hard samples since it further reduces the loss for the optimization formulation. 
Learning tasks from samples with different difficulty at different stages of learning leads to better generalization across tasks as we have seen in Tables \ref{tab:hsm_clean} and \ref{tab:cs_clean}.

\subsection{Learning with Noisy Annotations} \label{sec:learn_noisy}
In previous sections, we have seen the performance of our method in an ideal setup when there is no annotation noise, 
but often in real world datasets, we can expect to see some amount of noise or corruption in annotations. For multi-task 
learning, every instance has multiple annotations, one for each task. It is possible that one or more such annotations
can be corrupt. In such a situation, an ideal method must identify these corruptions and learn only from clean labels. 
In this section, we demonstrate the resilience of our method in the presence of noisy or corrupt labels.

\begin{figure*}[!t]
   \captionsetup{font=small}
   \begin{floatrow}
     \ffigbox{\includegraphics[scale=0.25]{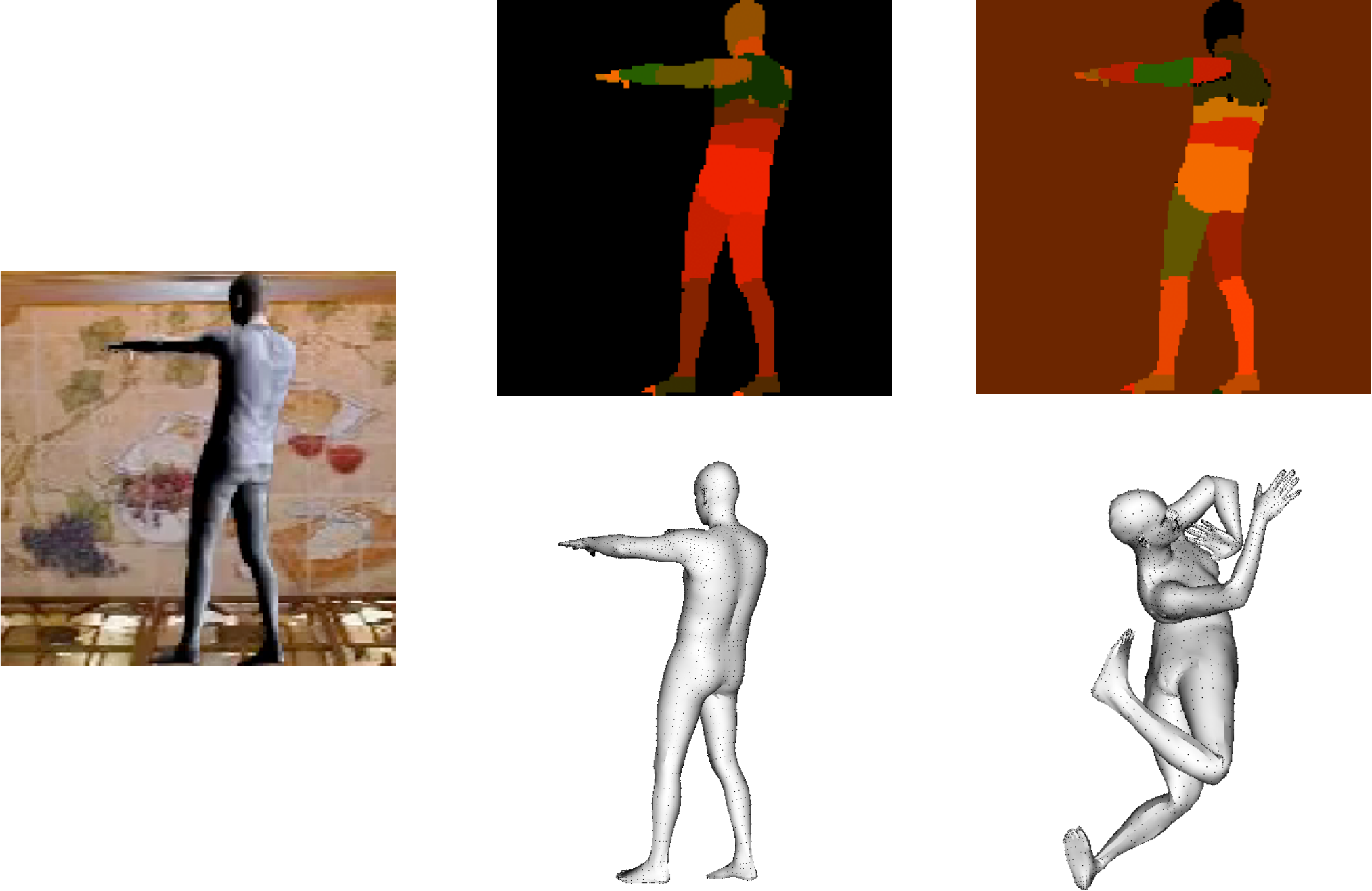}}
     {
     \caption{\textbf{Left}: Image from SURREAL dataset, \textbf{Middle}: Ground truth annotations provided by the dataset, \textbf{Right}: Synthesized noisy annotations from clean annotations} \label{fig:surreal_noise_sim}
     }
     \ffigbox{\includegraphics[scale=0.27]{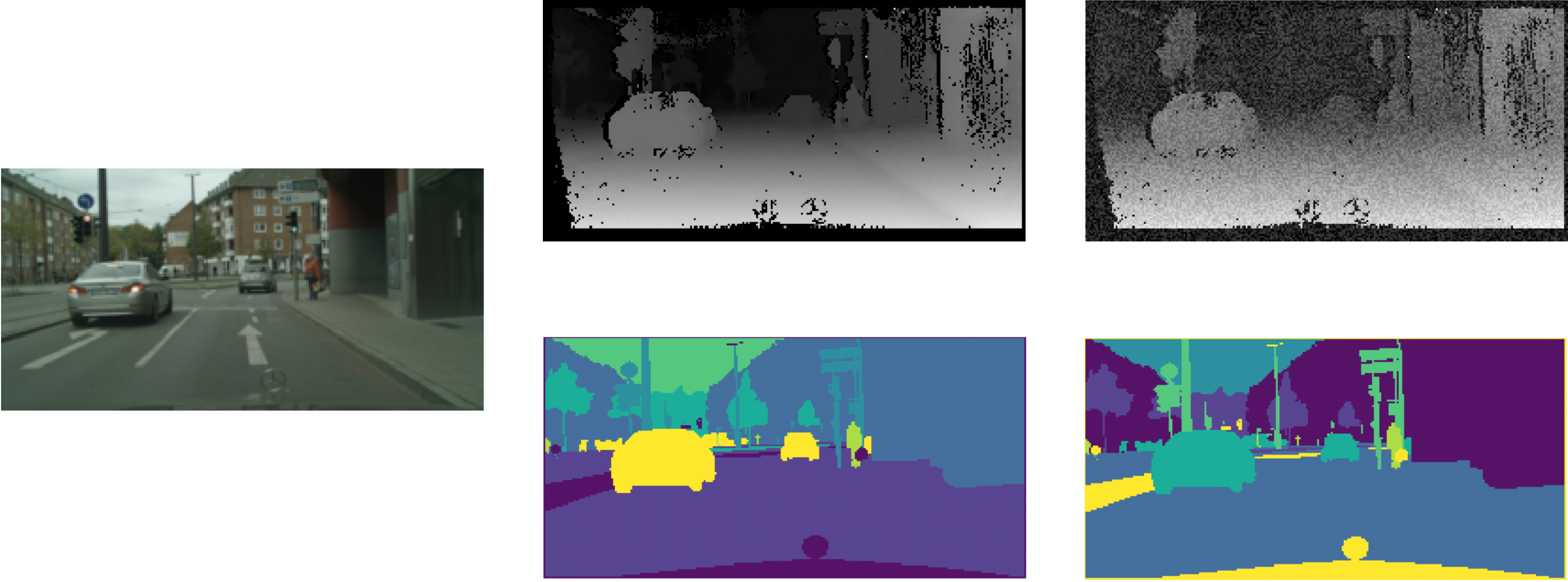}}
     {
     \caption{\textbf{Left}: Image from CityScapes dataset, \textbf{Middle}: Ground truth annotations provided by the dataset, \textbf{Right}: Synthesized noisy annotations from clean annotations} \label{fig:cs_noise_sim}
     }
   \end{floatrow}
\end{figure*}

\begin{table*}[!b]
\captionsetup{font=small}
\noindent
\begin{floatrow}\CenterFloatBoxes
\capbtabbox{
  \begin{adjustbox}{width=7.75cm}
  \begin{tabular}{ccccc} 
    \toprule
    \multirow{2}{*}[-3pt]{Method} & \multicolumn{2}{c}{Segmentation} & \multirow{2}{*}[5.3pt]{Depth} \\ 
    \cmidrule{2-4} & mIoU & Pix Acc & Abs Err \\ 
    \midrule
    Eq. Weight & 62.77 $\pm$ 0.17 & 94.15 $\pm$ 0.14 & 0.0266 $\pm$ 0.0026  & \\
    MTU~\cite{kendall2017multitask} & 62.36 $\pm$ 0.22 & 93.96 $\pm$ 0.07 & 0.0287 $\pm$ 0.0023 & \\
    DWA~\cite{liu2018endtoend} & 62.11 $\pm$ 0.80 & 93.65 $\pm$ 0.45 & 0.0374 $\pm$ 0.0044&  \\
    GLS~\cite{chennupati2019multinet} & 61.61 $\pm$ 0.27 & 93.67 $\pm$ 0.22 & 0.0328 $\pm$ 0.0022 & \\
    ILT Params (Ours) & \textbf{63.46 $\pm$ 0.03} & \textbf{94.28 $\pm$ 0.11} & \textbf{0.0223 $\pm$ 0.0004} & \\
    \bottomrule
  \end{tabular}
  \end{adjustbox}
 }{
  \caption{Comparison with baseline loss weighting methods on CityScapes validation dataset when 40\% of training dataset has corrupt depth labels.} \label{tab:cs_corruption_exps_depth}
}
\capbtabbox{
  \begin{adjustbox}{width=7.75cm}
  \begin{tabular}{ccccc} 
    \toprule
    \multirow{2}{*}[-3pt]{Method} & \multicolumn{2}{c}{Segmentation} & \multirow{2}{*}[5.3pt]{Depth} \\
    \cmidrule{2-4} & mIoU & Pix Acc & Abs Err \\ 
    \midrule
    Eq. Weight & 42.19 $\pm$ 0.66 & 78.99 $\pm$ 1.36 & 0.0193 $\pm$ 0.0050 & \\
    MTU~\cite{kendall2017multitask} & 47.04 $\pm$ 1.94 &  85.62 $\pm$ 2.01 & 0.0185 $\pm$ 0.0057 & \\
    DWA~\cite{liu2018endtoend} & 41.55 $\pm$ 2.02 & 78.30 $\pm$ 3.56 & 0.1949 $\pm$ 0.2525 &  \\
    GLS~\cite{chennupati2019multinet} & 46.95 $\pm$ 0.73 & 85.56 $\pm$ 1.13 & 0.0182 $\pm$ 0.0025 &  \\
    ILT Params (Ours) & \textbf{47.53 $\pm$ 0.87} & \textbf{86.87 $\pm$ 0.82} & \textbf{0.0123 $\pm$ 0.0005} & \\
    \bottomrule
  \end{tabular}
  \end{adjustbox}
 }{
  \caption{Comparison with baseline loss weighting methods on CityScapes validation dataset when 40\% of training dataset has corrupt segmentation labels.} \label{tab:cs_corruption_exps_segm}
}
\end{floatrow}
\end{table*}

\begin{figure*}[!t]
   \captionsetup{font=small}
     \ffigbox{\includegraphics[scale=0.335]{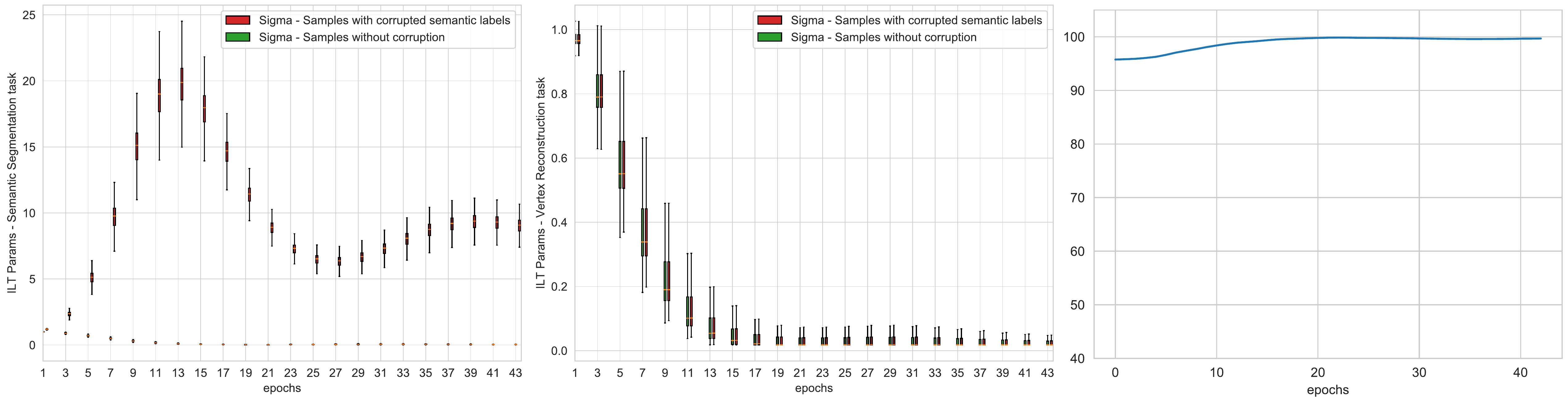}}
     {
     \caption{Comparison of instance-level task parameters for samples with noisy and clean segmentation labels \textbf{Left}: Instance-level task parameters for segmentation task, \textbf{Middle}: Instance-level task parameters for vertex reconstruction task, \textbf{Right}: Percentage of corrupt samples in the top 40\% data-points sorted by their instance parameter value for segmentation task.} \label{fig:sigma_analysis}
     }
\end{figure*}

\textbf{Controlled annotation corruption}: To study the effectiveness of our method in the presence of label noise, 
we simulate noise and corruption in CityScapes~\cite{cordts2016cityscapes} and SURREAL~\cite{surreal2017} datasets.
In SURREAL, to simulate corruption of semantic segmentation labels, we create a random permutation of human part labels.
We add uniform noise in the range -2.0 to 2.0 for pose and shape annotations, which results in the corruption shown in Figure \ref{fig:surreal_noise_sim}. We obtained the range by identifying the minimum and maximum values for pose and shape labels in the entire training dataset. Similarly in CityScapes, we add uniform noise in the range 0.0 to 0.2 to disparity labels. We obtained the range by identifying the minimum and maximum disparity values in the entire training dataset. To simulate corruption of semantic segmentation labels, we create a random permutation of semantic labels in the scene, see Figure \ref{fig:cs_noise_sim}.
Corruption is applied to a subset of training dataset before training and kept fixed 
for the entire training run. No corruption is applied to validation dataset. 
We report metrics for annotation corruption where 40\% of training dataset labels are corrupted. 
Ablative experiments on different levels of corruption are discussed in Section C of supplementary material. 

For DWA~\cite{liu2018endtoend}, we retain loss scaling values described in previous sections and search for the best temperature value. For experiments in Tables \ref{tab:hsm_corruption_exps_segm} and \ref{tab:hsm_corruption_exps_pose}, we found temperatures 1.8 and 0.5 to work the best respectively. We set learning rate for instance-level task parameters to 2.0 for both experiments. For experiments in Tables \ref{tab:cs_corruption_exps_depth} and \ref{tab:cs_corruption_exps_segm}, DWA temperature of 2.0 worked best for both experiments. Learning rate for instance-level task parameters was set to 0.4 and 3.0 respectively.
From Tables \ref{tab:hsm_corruption_exps_segm}, \ref{tab:hsm_corruption_exps_pose}, \ref{tab:cs_corruption_exps_depth} and \ref{tab:cs_corruption_exps_segm}, it is evident that our method performs well under high levels of label corruption. In Table \ref{tab:hsm_corruption_exps_pose}, we observe that ILT Params has 60\% lower surface errors compared to the second best multi-task loss weighting strategy MTU~\cite{kendall2017multitask}.
It is also interesting to note that our method has lower standard deviation for various metrics across multiple runs compared to other loss weighting schemes. Our method benefits from modeling the uncertainty independently for every instance and task. Hence for samples with corrupted labels for a task, the learning process can choose not to learn that task from those samples and pay a penalty of $\log(\sigma^2)$ while retaining the performance for tasks with clean labels. In the following section, we show how to exploit this property to identify corrupted labels in a dataset.

\textbf{Identifying corruption with instance-level task parameters}: 
We analyze instance-level task parameters from an experiment with 40\% label corruption, only for segmentation 
annotations of SURREAL dataset (shape and pose labels are accurate). 
We observe the behavior of parameters $\sigma^{cls}$ and $\sigma^{3d}$. We plot the distribution of $\sigma^{cls}$ 
and $\sigma^{3d}$ parameters at the end of every epoch of training, see left and middle plots of Figure \ref{fig:sigma_analysis}. 
We plot instance-level task parameters for samples with corrupted labels 
and compare them against samples without corruption. It is evident from the left and middle plots of Figure \ref{fig:sigma_analysis}
that our method chose to ignore learning segmentation task from samples with corrupted labels very early in the training process. Segmentation task parameters $\sigma^{cls}$ for noisy samples attain a very large value, which means the training process was willing to pay a penalty of $\log(\sigma^{cls})^2$. But for the same samples, it chose to learn vertex reconstruction task, as the box plots of $\sigma^{3d}$ for samples with clean and samples with corrupted segmentation labels are highly similar. In Figure \ref{fig:sigma_analysis}, we plot the percentage of corrupted samples in top 40\% of training data sorted by their $\sigma^{cls}$ values. As seen from the plot, by 20$^{th}$ epoch, 99.7\% of samples with noisy segmentation labels attain values greater than samples with clean segmentation labels. Similar trend is observed when pose and shape labels are corrupted instead of segmentation labels. In Table \ref{tab:hsm_cs_noise_detection} we report accuracies of identifying samples with corrupt labels for different tasks. 
Accuracies are evaluated by simply sorting instance-level task parameters in decreasing order of maginitude for the task that has corrupt labels and identifying how many samples with corrupt labels were found in top 20\% and 40\% of training data for 20\% and 40\% label corruption. From Table \ref{tab:hsm_cs_noise_detection}, we see for various levels of corruption across tasks in SURREAL and CityScapes, our method can effectively identify samples with corrupt labels.

\begin{table}[!t]
\captionsetup{font=small}
  \begin{adjustbox}{width=7.8cm}
  \begin{tabular}{cccc} 
    \toprule 
    Dataset & Corrupt Label & \% Corrupt & Detection Accuracy \\
    \midrule
    SURREAL & Pose \& Shape & 20\% & 97.79\% \\
     & & 40\% & 98.99\% \\
    \cmidrule(lr){2-4}
     & Segmentation & 20\% & 99.85\% \\
     & & 40\% & 99.79\% \\
    \midrule
    \midrule
    CityScapes & Depth & 20\% & 100\% \\
     & & 40\% & 100\% \\
    \cmidrule(lr){2-4}
    & Segmentation & 20\% & 97.14\% \\
     & & 40\% & 96.01\% \\

    \bottomrule
    \end{tabular}
  \end{adjustbox}
{
  \caption{Percent of samples with corrupt labels found in top 20\% and top 40\% of SURREAL and CityScapes training datasets when sorted by magnitude of instance-level task parameters for 20\% and 40\% label corruption respectively.} \label{tab:hsm_cs_noise_detection}
}
\end{table}

\section{Conclusion}
In this paper, we introduced instance-level task parameters, where every sample in the training set is equipped with a set of learnable parameters, where the cardinality is equal to the number of tasks learned by the model. These parameters model the uncertainty of a task for every sample, thus break the assumption made by prior works that all samples are equally hard for a task. Our method prioritizes learning from samples with clean labels in the presence of corruption, and they can be used to identify samples with corrupted labels post training. Moreover, instance-level task parameters are only utilized in training, hence they do not affect the model capacity and they are not used at inference. We have conducted experiments on SURREAL and CityScapes datasets and showed that our method performs well against recent multi-task learning baselines in both clean and corrupt settings.

{\small
\bibliographystyle{ieee_fullname}
\bibliography{egbib}
}

\end{document}